\documentclass[a4paper]{article}
\usepackage{fip}
\usepackage{graphicx}
\usepackage{times}
\usepackage{mathptm}

\newtheorem{definition}{Definition}
\begin{document}
\title{
\bf\Large A NEW VALIDITY MEASURE FOR FUZZY C-MEANS CLUSTERING
\thanks {This work was supported by the Korea Science and
Engineering Foundation (KOSEF)  through the Advanced Information
Technology Research Center} }
\author{\normalsize
         Dae-Won Kim$^{1}$\quad
         Kwang H. Lee$^{1,2}$\\
         \normalsize\sl
         1.Department of Computer Science, KAIST, Daejon, Korea\\
         \normalsize\sl
         2.Department of BioSystems, KAIST, Daejon, Korea \\
         \normalsize\sl
         Email:dwkim@if.kaist.ac.kr
        }
\date{}
\maketitle

\begin{abstract}
A new cluster validity index is proposed for fuzzy clusters
obtained from fuzzy $c$-means algorithm. The proposed validity
index exploits inter-cluster proximity between fuzzy clusters.
Inter-cluster proximity is used to measure the degree of overlap
between clusters. A low proximity value refers to well-partitioned
clusters. The best fuzzy $c$-partition is obtained by minimizing
inter-cluster proximity with respect to $c$. Well-known data sets
are tested to show the effectiveness and reliability of the
proposed index.
\end{abstract}

{\bf Keyword: }Cluster validity, Fuzzy clustering, Fuzzy c-means

\section{\normalsize INTRODUCTION}

The objective of fuzzy clustering is to partition a data set into
$c$ homogeneous fuzzy clusters. The most widely used fuzzy
clustering algorithm is the Fuzzy C-Means (FCM) algorithm proposed
by Bezdek~\cite{Bez81}. The algorithm, however, requires a
pre-defined number of clusters ($c$) by a user. Different
selections on the initial number of clusters result in different
clustering partitions. Thus, it is necessary to validate each of
the fuzzy c-partitions once they are found. This evaluation is
called cluster validity. Many fuzzy cluster validity indexes have
been proposed in the
literature~\cite{Bez81}\cite{Bez74a}\cite{Bez74b}\cite{Pal95}\cite{Xie91}\cite{Fuk89}\cite{Kwo98}\cite{Rez98}\cite{Bou99}.
Bezdek's partition coefficient (PC)~\cite{Bez74a}, partition
entropy (PE)~\cite{Bez74b}, and Xie-Beni's index~\cite{Xie91} have
been frequently used in current research. In this paper, a new
cluster validity index for fuzzy clustering is proposed. The
proposed index evaluates the fuzzy cluster partitions based on a
proximity relation between fuzzy clusters. The proximity value is
both the degree of overlapping and the inverse distance of
separation between the clusters. Thus, a smaller proximity value
indicates a lower degree of overlapping and farther separation
between the clusters, which results in well-partitioned fuzzy
clusters.

\section{\normalsize PREVIOUS WORK}

The FCM clustering algorithm has been widely used to obtain the
fuzzy $c$-partition. However, the algorithm may fall into local
optima due to the initial selection of cluster centroids. Most of
the clustering algorithms take a random generation to the initial
selection method, which result in the fact that different
selections on the initial cluster centroids cause different
clustering partitions. Thus, an evaluation methodology was
required to validate each of the fuzzy c-partitions once they are
found. This evaluation is called cluster validity. In addition, a
cluster validity can help to find out the optimal number of
clusters ($c$) when the number of clusters is not known in a
priori~\cite{Rez98}.

The objective of the FCM is to obtain the fuzzy $c$-partition
$\tilde{F}=\{ \tilde{F}_1,\tilde{F}_2,..,\tilde{F}_c\}$ for both
an unlabeled data set $X=\{ x_1, ...,x_n\}$ and the number of
clusters $c$ by minimizing the evaluation function $J_m$:

\begin{equation}
minimize~
J_m(U,V:X)=\sum_{i=1}^{c}\sum_{j=1}^{n}(\mu_{ij})^{m}\|x_j-v_i\|^2
\end{equation}

where $\mu_{ij}$ is the membership degree of data $x_j$ to a fuzzy
cluster set $\tilde{F}_i$, and also, is an element of a $(c \times
n)$ pattern matrix $U=[\mu_{ij}]$. $V=(v_1,v_2,..,v_c)$ is a
vector of cluster centers of the fuzzy $c$-partition $\tilde{F}$.
$\|x_j-v_i\|^2$ is an Euclidean norm between $x_j$ and $v_i$. The
parameter $m$ controls the fuzziness of membership of each data.
Most of fuzzy clustering algorithms are performed using $m=2$
because it gives better performance results than other $m$
values~\cite{Pal95}.

Bezkek proposed two cluster validity indexes for fuzzy
clustering~\cite{Bez74a}\cite{Bez74b}: Partition Coefficient
($V_{PC}$) and Partition Entropy ($V_{PE}$). Equation~\ref{eq:pc}
shows that $V_{PC}$ takes its maximum as an optimal partition, and
$V_{PE}$ takes its unique minimum as the best one.
\begin{equation}\label{eq:pc}
V_{PC} = \frac{\sum_{j=1}^{n}\sum_{i=1}^{c}\mu_{ij}^{2}}{n},
\qquad V_{PE} = -
\frac{1}{n}\sum_{j=1}^{n}\sum_{i=1}^{c}[\mu_{ij}~log_{a}(\mu_{ij})]
\end{equation}

Xie and Beni proposed a validity index $V_{XB}$ that focused on
two concepts: compactness and separation~\cite{Xie91}. The
numerator part of equation~\ref{eq:xb} indicates the compactness
of fuzzy partition. The denominator part explains the level of
separation between clusters. Fukuyama and Sugeno also tried to
model the cluster validation $V_{FS}$ by exploiting both the
compactness of cluster and the distance from each cluster centroid
in equation~\ref{eq:fs}~\cite{Fuk89}. Kwon extended the Xie and
Beni's index to eliminate its monotonic decreasing
tendency~\cite{Kwo98}.
\begin{equation}\label{eq:xb}
V_{XB} =
\frac{\sum_{i=1}^{c}\sum_{j=1}^{n}\mu_{ij}^{2}\|x_j-v_i\|^{2}}{n(\underbrace{min}_{i
\neq k}\|v_i-v_k\|^2)}
\end{equation}
\begin{equation}\label{eq:fs}
V_{FS} =
\sum_{j=1}^{n}\sum_{i=1}^{c}\mu_{ij}^{2}(\|x_k-v_i\|^{2}-\|v_i-
\bar{v}\|^{2})
\end{equation}
\begin{equation}
V_{K}
=\frac{\sum_{j=1}^{n}\sum_{i=1}^{c}\mu_{ij}^{2}\|x_j-v_i\|^{2}+\frac{1}{c}\sum_{i=1}^{c}\|v_i-\bar{v}\|^{2}}{\underbrace{min}_{i
\neq k}\|v_i-v_k\|^2}
\end{equation}

More recent indexes in cluster validation are associated with the
degree of variance in each cluster. Rezaee combined the
within-cluster variance with the distance functional~\cite{Rez98}.
In equation~\ref{eq:cwb}, $V_{CWB}$, $\sigma(v_i)$ indicates the
fuzzy variance of the $i$-th cluster, and $\sigma(X)$ means the
variance of the pattern set X. Boudraa also suggested to take the
cluster variance into consideration for the validity index
$V_{B_{crit}}$ ~\cite{Bou99}.
\begin{equation}\label{eq:cwb}
V_{CWB} = \alpha
\frac{\sum_{i=1}^{c}\|\sigma(v_i)\|}{c\|\sigma(X)\|} +
\frac{D_{max}}{D_{min}}\sum_{k=1}^{c}(\sum_{z=1}^{c}\|v_k-v_z\|)^{-1}
\end{equation}
\begin{equation}
V_{B_{crit}} = \frac{max~\delta(v_i,v_j)}{\underbrace{min}_{i \neq
j}~\delta(v_i,v_j)} + \alpha
\frac{1}{c}\frac{\sum_{q=1}^{P}\sum_{k=1}^{c}var_{q}(k)}{\sum_{q=1}^{P}var_{t}(q)}
\end{equation}

\section{\normalsize PROPOSED CLUSTER VALIDITY MEASURE}

\subsection{\normalsize Motivation and intuition}

In this paper, a new cluster validity index for the FCM is
proposed. The key concept is to exploit the geometric properties
of fuzzy clusters. All the previous indexes including $V_{PC}$ and
$V_{PE}$ focused on only the compactness and the variation of
within-cluster. Some indexes such as $V_{XB}$ and $V_{CWB}$ have
used the strength of separation between clusters in order to
account for an inter-cluster structural information. However,
those indexes are limited in their ability to provide a meaningful
interpretation of structure in the data since the separation is
simply computed by considering only the distance between cluster
centroids.

\begin{figure}[t]
\begin{center}
\includegraphics[width=8cm]{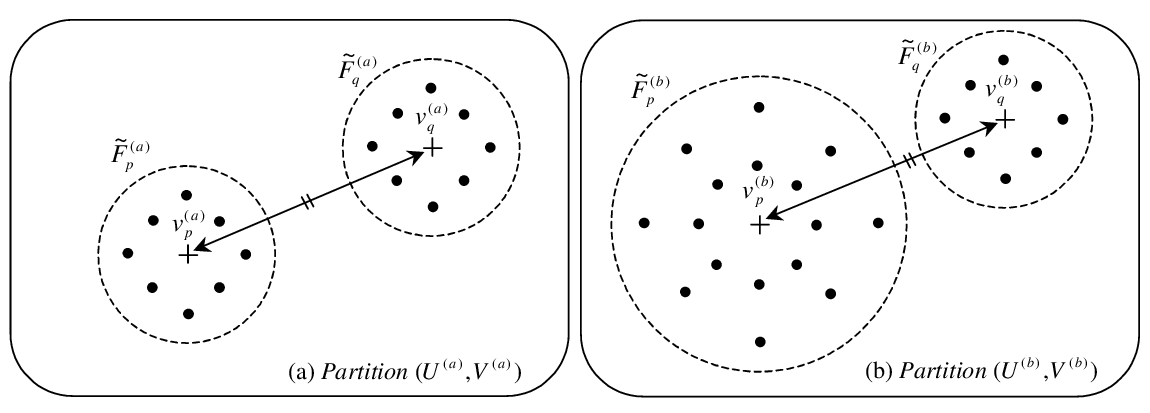}
\end{center}
\caption{Two different fuzzy partitions $(U^{(a)},V^{(a)})$ and
$(U^{(b)},V^{(b)})$ containing the same separation distance
between cluster centroids} \label{fig:wrong-separation}
\end{figure}

Figure~\ref{fig:wrong-separation} shows an example of how the
traditional indexes are limited in their ability to recognize the
correct difference between clusters. There are two different fuzzy
partitions $(U^{(a)},V^{(a)})$ and $(U^{(b)},V^{(b)})$ containing
the same separation distance between cluster centroids. Two fuzzy
clusters $\tilde{F}_{p}^{(a)}, \tilde{F}_{q}^{(a)} \in U^{(a)}$
have their centroids $v_{p}^{(a)}$, $v_{q}^{(a)} \in V^{(a)}$
respectively in figure~\ref{fig:wrong-separation}(a). In
figure~\ref{fig:wrong-separation}(b), two fuzzy clusters
$\tilde{F}_{p}^{(b)}, \tilde{F}_{q}^{(b)} \in U^{(b)}$ have their
own centroids $v_{p}^{(b)}$,$v_{q}^{(b)} \in V^{(b)}$,
respectively. We find that $(U^{(a)},V^{(a)})$ is better separated
than $(U^{(b )},V^{(b)})$ because a large number of data $x_j \in
X$ are more clearly classified in $(U^{(a)},V^{(a)})$. However,
with the conventional indexes, it is difficult to tell which
partition of the two is better in viewpoints of isolation and hard
to differentiate the geometric structure between two partitions
since the separation distances $\|v_{p}^{(a)}-v_{q}^{(a)}\|$ and
$\|v_{p}^{(b)}-v_{q}^{(b)}\|$ are equal. The limitation is due to
the simple measure of separation between cluster centroids.

Thus it is necessary to devise a new validity index that is
capable of utilizing the overall geometric structure between
clusters and deriving a sufficient interpretation from the
structure in the data. We take a novel approach based on an
inter-cluster proximity measure between fuzzy clusters. Each fuzzy
cluster is represented and regarded as a fuzzy set in
equation~\ref{eq:fuzzyset}.

\begin{equation}\label{eq:fuzzyset}
\tilde{F}_i=\sum_{j=1}^{n}\mu_{\tilde{F}_i}(x_j)/x_j=\mu_{\tilde{F}_i}(x_1)/x_1+
\mu_{\tilde{F}_i}(x_2)/x_2+...+\mu_{\tilde{F}_i}(x_n)/x_n
\end{equation}

Given fuzzy partition $(U,V)$, the proposed index attempts to
obtain the $proximity$ between fuzzy clusters by computing the
similarity between the fuzzy cluster sets. The inter-cluster
proximity indicates both the degree of overlapping and the inverse
distance of separation between the clusters. In other words, a
smaller proximity value means a lower degree of overlapping and
farther separation between the clusters, which results in
well-partitioned fuzzy clusters.

\subsection{\normalsize Validity index using inter-cluster proximity}

With each fuzzy cluster
$\tilde{F}_i=\sum_{j=1}^{n}\mu_{\tilde{F}_i}(x_j)/x_j=\mu_{\tilde{F}_i}(x_1)/x_1+
\mu_{\tilde{F}_i}(x_2)/x_2+...+\mu_{\tilde{F}_i}(x_n)/x_n$, we
obtain the proximity value between two fuzzy clusters at each
point of membership degree ($\mu$) before computing the total
inter-cluster proximity. The proximity function $f(\mu)$ at a
given membership degree $\mu$ between two fuzzy clusters
$\tilde{F}_p$ and $\tilde{F}_q$ is defined as:

\begin{equation}\label{eq:fu}
f(\mu:\tilde{F}_p,\tilde{F}_q)=\sum_{j=1}^{n}\delta(x_j,\mu:\tilde{F}_p,\tilde{F}_q)\omega(x_j)
\end{equation}

where

\begin{equation}\label{eq:delta}
\delta(x_j,\mu:\tilde{F}_p,\tilde{F}_q)= \left \{
\begin{array}{ll}
1.0 & \quad \mbox{if $\mu \leq MIN(\mu_{\tilde{F}_p}(x_j),\mu_{\tilde{F}_q}(x_j))$} \\
0.0 & \quad \mbox{otherwise}
\end{array}
\right.
\end{equation}

$\delta(x_j,\mu:\tilde{F}_p,\tilde{F}_q)$ determines whether two
clusters are proximate at the membership degree $\mu$ for data
$x_j$. It returns that proximity is 1.0 when both membership
degrees of the two clusters are more than $\mu$; otherwise, it
returns 0.0. Furthermore, $\omega(x_j)$ is introduced to assign a
weight for vague data. The weight value of $\omega(x_j) \in
[0.0,1.0]$ is determined by the degree of sharing of data $x_j$
between clusters. Vague data are given more weight than clearly
classified data. For example, $\omega(x_j)$ is given a value
within $[0.7,1.0]$ when $x_j$ is considered vague such that
$\mu_{\tilde{F}_i}(x_j) \leq 0.5$ for all $\tilde{F}_i \in
\tilde{F}$. Conversely, $\omega(x_j)$ is given a weight within
$[0.0,0.3]$ when $x_j$ is clearly classified as belonging to one
of the clusters such that $\mu_{\tilde{F}_i}(x_j) \geq 0.8$ for
any $\tilde{F}_i \in \tilde{F}$. This approach makes it possible
for us to focus more concentration on the highly-overlapped vague
data in the computation of the validity index than other indexes
do.

\begin{figure}[t]
\begin{center}
\includegraphics[width=12cm]{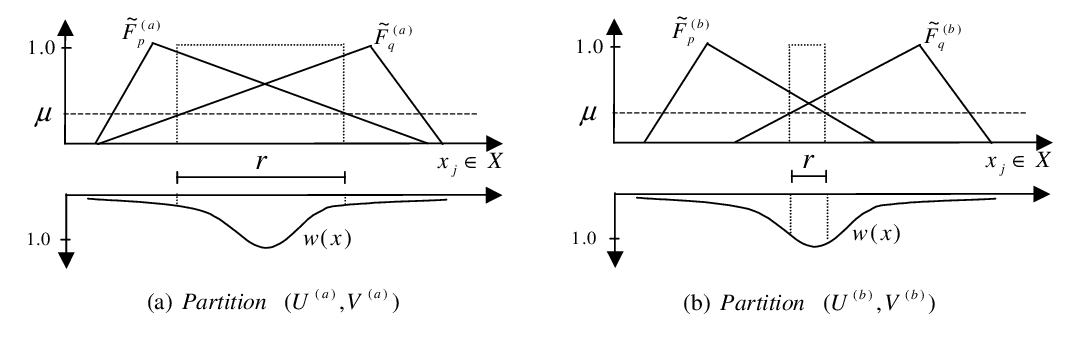}
\end{center}
\caption{Proximity values $f(\mu)$ at membership degree $\mu$
between two fuzzy clusters} \label{fig:proximity}
\end{figure}

Figure~\ref{fig:proximity} represents two proximity values
$f(\mu)$ at membership degree $\mu$ between two fuzzy clusters.
Two fuzzy partitions $U^{(a)}$ and $U^{(b)}$ were shown in the
figure. In the figure~\ref{fig:proximity}(a), among all $x_j \in
X$, only $x_j \in r$ are given a value 1.0 by
equation~\ref{eq:delta}. Given the weight function $w(x)$,
proximity $f(\mu:\tilde{F}_p^{(a)},\tilde{F}_q^{(a)})$ is obtained
by aggregating the product of
$\delta(x_j,\mu:\tilde{F}_p^{(a)},\tilde{F}_q^{(a)})$ and $w(x_j)$
for all $x_j \in X$ by equation~\ref{eq:fu}. When we compare two
proximity values $f(\mu)$ for the partitions $U^{(a)}$ and
$U^{(b)}$, the $f(\mu:\tilde{F}_p^{(a)},\tilde{F}_q^{(a)})$ is
given more higher value than
$f(\mu:\tilde{F}_p^{(b)},\tilde{F}_q^{(b)})$. The number of
overlapped data $x_j \in r$ in the partition $U^{(a)}$ is more
than in $U^{(b)}$, which indicates that the partition $U^{(a)}$
has much more vague data than $U^{(b)}$. Thus, we can see that the
partition $U^{(b)}$ with a lower $f(\mu)$ is a better partition
than $U^{(a)}$ with a higher $f(\mu)$.

\begin{definition}[Inter-cluster proximity]\label{def:proximity}
Let $\tilde{F}_p$ and $\tilde{F}_q$ be two fuzzy clusters
belonging to a pattern matrix $U$. Let
$f(\mu:\tilde{F}_p,\tilde{F}_q)$ be a proximity function at a
given membership degree $\mu$ between $\tilde{F}_p$ and
$\tilde{F}_q$. Then, a total inter-cluster proximity
$S(\tilde{F}_p,\tilde{F}_q)$ between $\tilde{F}_p$ and
$\tilde{F}_q$ is defined as
\begin{equation}
S(\tilde{F}_p,\tilde{F}_q) =
\sum_{\mu}f(\mu:\tilde{F}_p,\tilde{F}_q) =
\sum_{\mu}\sum_{j=1}^{n}\delta(x_j,\mu:\tilde{F}_p,\tilde{F}_q)\omega(x_j)
\end{equation}
\end{definition}

$S(\tilde{F}_p,\tilde{F}_q)$ is obtained by calculating the
$f(\mu:\tilde{F}_p,\tilde{F}_q)$ for the whole range of membership
degrees. A small value for $S(\tilde{F}_p,\tilde{F}_q)$ indicates
that $\tilde{F}_p$ has little proximity to $\tilde{F}_q$, which
results in a well-separated partition between the two clusters.
With the above proximity definition for the two fuzzy clusters, we
establish a new cluster validity index for fuzzy clustering.

\begin{definition}[Proposed validity index]\label{def:proposed-index}
Let $\tilde{F}_p$ and $\tilde{F}_q$ be two fuzzy clusters
belonging to a pattern matrix $U$. Let
$S(\tilde{F}_p,\tilde{F}_q)$ be the inter-cluster proximity
between $\tilde{F}_p$ and $\tilde{F}_q$. Let the number of
clusters and the computation number for the proximity between the
fuzzy clusters be denoted by $c$ and $_cC_{2}$, respectively.
Then, the proposed validity index $V_{proposed}(U,V:X)$ is defined
as
\begin{equation}
\begin{array}{l} V_{proposed}(U,V:X) =
\frac{1}{_cC_{2}}\underbrace{\sum_{p=1}^{c}\sum_{q=1}^{c}}_{p \neq
q }S(\tilde{F}_p,\tilde{F}_q) \\ \\ \qquad\qquad\qquad\qquad=
\frac{2}{c(c-1)}\underbrace{\sum_{p=1}^{c}\sum_{q=1}^{c}}_{p \neq
q
}[\sum_{\mu}\sum_{j=1}^{n}\delta(x_j,\mu:\tilde{F}_p,\tilde{F}_q)\omega(x_j)]
\end{array}
\end{equation}
\end{definition}

$V_{proposed}$ index takes an average proximity for all clusters
of fuzzy $c$-partition. We see that a small value for
$V_{proposed}$ indicates a well-clustered fuzzy $c$-partition as
suggested by Definition~\ref{def:proposed-index}. Therefore, the
best fuzzy $c$-partition or the optimum value of $c$ is obtained
by minimizing $V_{proposed}$ over $c=2,3,...,c_{max}$.

\section{\normalsize EXPERIMENTAL RESULTS}

To show the reliability and effectiveness of the proposed index in
finding the optimal partition, extensive comparisons with other
indexes are conducted on five widely used data sets. The proposed
index $V_{proposed}$ was compared with seven fuzzy cluster
validity indexes for the fuzzy $c$-partition $(U,V)$ obtained from
the FCM: $V_{PC}$~\cite{Bez74a}, $V_{PE}$~\cite{Bez74b},
$V_{XB}$~\cite{Xie91}, $V_{FS}$~\cite{Fuk89}, $V_K$~\cite{Kwo98},
$V_{CWB}$~\cite{Rez98}, $V_{B_{crit}}$~\cite{Bou99}. For each data
set, we made several runs of the FCM for different values of
$c=2,..,c_{max}$. The parameters of the FCM were set as follows:
termination criterion $\epsilon=0.001$, the weighting exponent
$m=2.0$, and $\|\ast\|^2$ was a square of the Euclidean norm.
Initial centroids $V_{0}$ were randomly selected.

\begin{figure}[t]
\begin{center}
\includegraphics[width=12cm]{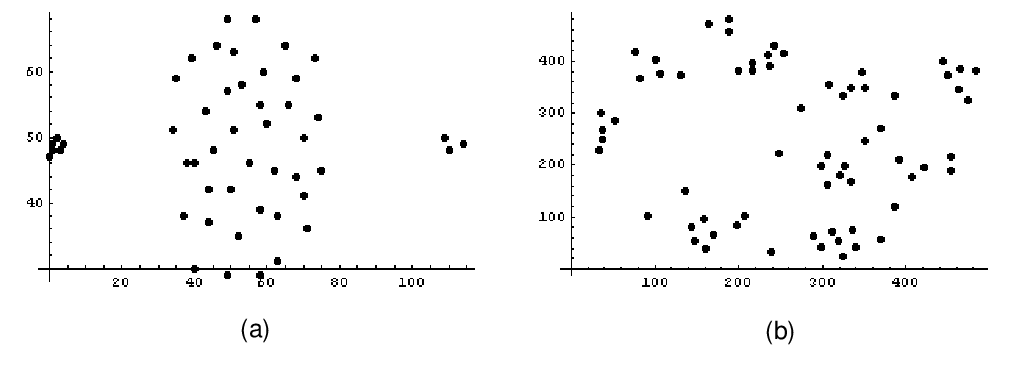}
\end{center}
\caption{(a) BENSAID data set (optimal $c$ is three) (b) STARFIELD
data set (optimal $c$ is eight) } \label{fig:dataset}
\end{figure}

\begin{table}[t]
\label{table:bensaid-result}
\begin{center}
\caption{Cluster validity values on the BENSAID data set for
$c=2,..,c_{max}=\sqrt{n} \approx 7$}
\begin{tabular}{ccccccccc}
\hline $c$ & $V_{PC}$ & $V_{PE}$ & $V_{XB}$ & $V_{FS}$ & $V_K$ & $V_{CWB}$ & $V_{B_{crit}}$ & $V_{proposed}$\\
\hline
$c=2$  & 0.72          & \textbf{0.19} & 0.24             & 3671.01               & 11.89            & 0.84             & 8.00             & 112.60 \\
$c=3$  & \textbf{0.75} & 0.20          & \textbf{0.07}     & -15676.31             & \textbf{4.12} & 0.62             & \textbf{4.46} & \textbf{39.87} \\
$c=4$  & 0.61          & 0.32          & 0.27             & -15035.68             & 15.87            & 0.58             & 8.19             & 80.90 \\
$c=5$  & 0.66          & 0.29          & 0.12             & -27285.22             & 8.71             & 0.46             & 7.69             & 58.96 \\
$c=6$  & 0.63          & 0.33          & 0.10             & -28692.18             & 8.14             & \textbf{0.43}             & 8.15             & 55.59 \\
$c=7$  & 0.61          & 0.36          & 0.10             & \textbf{-29292.01} & 9.20                & 0.44 & 9.09             & 57.52 \\
\hline
\end{tabular}
\end{center}
\end{table}

Figure~\ref{fig:dataset}(a) scatterplots the first data set that
was described by Bendsaid~\cite{Ben96}. The data set includes 49
data points in two dimensional measurement space, and consists of
three clusters. Table 1 shows the results of validity indexes with
respect to $c=2,3,...,c_{max}=7$. The optimal value $c$ of each
index is marked by bold face. For each $c \geq 2$, eight validity
indexes compute their index values. As mentioned before, all
indexes except $V_{PC}$ take their minimums as optimal values. The
number of clusters $c = 3$ in the Bendsaid's set was correctly
recognized by five cluster validity indexes including the proposed
index. $V_{PE}$ indicates the presence of two clusters. $V_{FS}$
and $V_{CWB}$ take their minimum at $c=7$ and $c=6$, respectively.

\begin{table}[b]
\label{table:starfield-result}
\begin{center}
\caption{Cluster validity values on the STARFIELD data set for
$c=2,..,c_{max}=\sqrt{n} \approx 8$}
\begin{tabular}{ccccccccc}
\hline $c$ & $V_{PC}$   & $V_{PE}$ & $V_{XB}$ & $V_{FS}$ & $V_K$ & $V_{CWB}$ & $V_{B_{crit}}$ & $V_{proposed}$\\
\hline
$c=2$  & \textbf{0.73}  & \textbf{0.18}    & 0.24               & 216235.05                 & 16.04         & 0.17             & 4.90               & 148.60 \\
$c=3$  & 0.66           & 0.26             & 0.12               & -597582.27                & 8.29          & 0.12             & 4.23               & 93.33 \\
$c=4$  & 0.62           & 0.32             & 0.12               & -835444.94                & 8.74          & 0.10             & 4.19               & 94.67 \\
$c=5$  & 0.63           & 0.33             & 0.11               & -1047072.42               & 8.16          & 0.09             & \textbf{4.09}      & 78.64 \\
$c=6$  & 0.65           & 0.33             & \textbf{0.10}      & -1266918.83               & \textbf{8.09} & 0.08             & 4.30               & 62.04 \\
$c=7$  & 0.66           & 0.33             & 0.11               & \textbf{-1394217.46}      & 9.61          & 0.07             & 4.66               & 57.83 \\
$c=8$  & 0.67           & 0.33             & 0.12               & -1368962.28               & 10.42         & \textbf{0.07}    & 5.10               & \textbf{50.59} \\
\hline
\end{tabular}
\end{center}
\end{table}

Figure~\ref{fig:dataset}(b) shows a superset of the STARFIELD data
set~\cite{Xie91}. The data set contains 66 data points where eight
or nine clusters are in agreement at a reasonably optimal
partition. Table 2 shows the results of validity indexes with
respect to $c=2,3,..,c_{max}=8$. We see that $V_{CWB}$ and
$V_{proposed}$ correctly recognize the presence of eight clusters.
The six remaining indexes do not agree with the optimal $c$ value.
$V_{PC}$ and $V_{PE}$ consider two clusters to be a natural
structure. The minimum values of $V_{XB}$ and $V_K$ is obtained at
$c=6$. $V_{FS}$ points to $c=7$. $V_{B_{crit}}$ indexes recognize
$c=5$ clusters as an optimal partition.

\begin{table}[t]
\label{table:summary}
\begin{center}
\caption{Preferable values of $c$ for six data sets by each
cluster validity index}
\begin{tabular}{cccccccccc}
\hline Data set & $c_{optimal}$ & $V_{PC}$ & $V_{PE}$ & $V_{XB}$ & $V_{FS}$ & $V_K$ & $V_{CWB}$ & $V_{B_{crit}}$ & $V_{proposed}$\\
\hline
BENSAID   & 3 & 3 & 2 & 3 & 7  & 3 & 7  & 3 & 3 \\
STARFIELD & 8 & 2 & 2 & 6 & 7  & 6 & 8  & 5 & 8 \\
IRIS      & 2 & 2 & 2 & 2 & 3  & 2 & 3  & 6 & 2 \\
X30       & 3 & 3 & 3 & 3 & 4  & 2 & 3  & 3 & 3 \\
BUTTERFLY & 2 & 2 & 2 & 2 & 3  & 2 & 2  & 2 & 2 \\
\hline
\end{tabular}
\end{center}
\end{table}

Table 3 summaries the comparison results of the validity indexes
of five well-known data sets: BENSAID's set~\cite{Ben96},
STARFIELD~\cite{Xie91}, IRIS~\cite{Pal95}, X30~\cite{Bez98}, and
BUTTERFLY~\cite{Kwo98}. IRIS has $n=150$ data points in a
four-dimensional measurement space. One can argue in favor of both
$c=2$ and $c=3$ for IRIS because of the substantial overlap of two
of the clusters. In this paper, we take $c=2$ as an optimal choice
in view of the geometric structure of IRIS~\cite{Pal95}. The X30
data set has $c=3$ compact, well-separated clusters with $n=30$
where each cluster has 10 points~\cite{Bez98}. The BUTTERFLY data
set contains $c=2$ clusters with $n=15$ data points~\cite{Kwo98}.
The column $c_{optimal}$ in Table 3 refers to the optimal number
of clusters for the given data sets, and the other columns show
the cluster numbers of each index. We find that $V_{proposed}$ is
the most reliable indexes since it correctly recognizes the number
of clusters for all data sets. On the other hand, $V_{PC}$,
$V_{PE}$, and $V_{XB}$ don't recognize the correct $c$ for
STARFIELD. $V_{PE}$ also fails to identify $c_{optimal}$ in
BENSAID set. We see that $V_{K}$ fails to recognize $c_{optimal}$
in the STARFIELD and X30 sets. The $V_{CWB}$ fails to recognize
the optimal number of clusters for BENSAID and IRIS sets. The
minimum value of $V_{B_{crit}}$ is obtained at $c = 5$ and $c=6$
in the STARTFIELD and IRIS set, respectively. $V_{FS}$ is much
more unreliable than the other indexes.

\section{\normalsize Conclusions}

In this paper, a new cluster validity index for the fuzzy
$c$-partition has been proposed. The validity index takes the
proximity between fuzzy clusters into account to evaluate the
fuzzy partitions. Given a fuzzy partition, the proposed index
computes an inter-cluster proximity based on a similarity in order
to obtain meaningful structural information. The best fuzzy
$c$-partition is obtained by minimizing the inter-cluster
proximity with respect to $c$. The performance of the proposed
index was tested on various data sets demonstrating its
effectiveness and reliability. From the experiments that find the
optimal number of clusters, the results indicate that the proposed
approach is more reliable than other indexes. In future work, we
plan to apply the proposed method to more realistic examples such
as color image clustering. We would like to obtain the optimal
clustering result of color images with the help of the validity
index.

\end{document}